\pdfoutput=1

\documentclass[11pt]{article}

\usepackage[]{emnlp}

\usepackage{times}
\usepackage{latexsym}

\usepackage[T1]{fontenc}

\usepackage[utf8]{inputenc}

\usepackage{microtype}

\usepackage{inconsolata}

\usepackage{booktabs}
\usepackage{graphicx}
\usepackage{multirow}
\usepackage{subcaption}
\usepackage{enumitem}
\usepackage{fdsymbol}
\usepackage{natbib}
\title{\includegraphics[width=16pt,height=16pt]{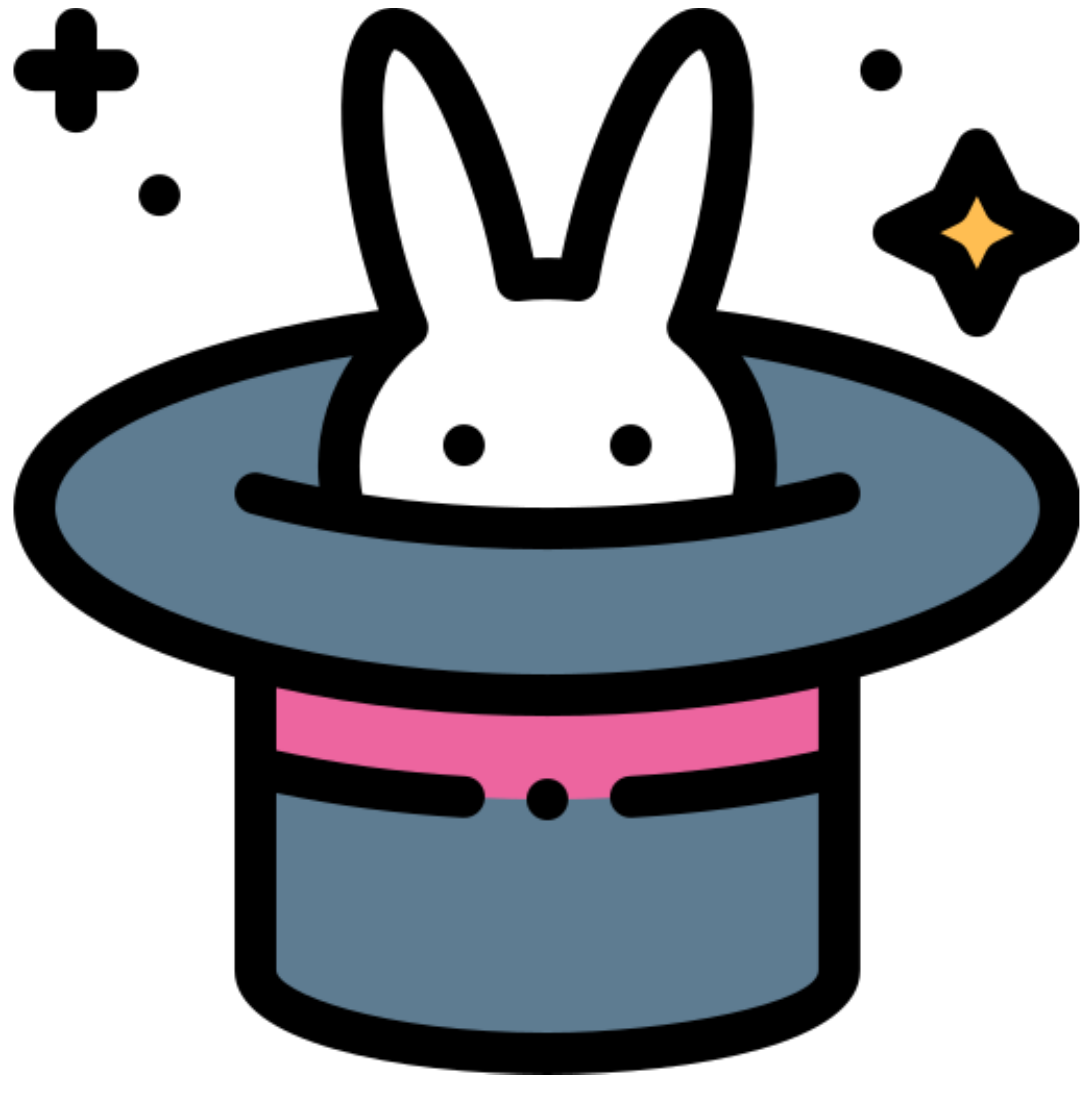}~MAGE: Machine-generated Text Detection in the Wild}

\author{ 
 Yafu Li$^{\spadesuit \clubsuit}\footnotemark[1]$\hspace{0.5mm},
 Qintong Li\hspace{0.5mm},
 Leyang Cui$^{\heartsuit}\footnotemark[2]$\hspace{0.5mm}, 
 Wei Bi$^{\heartsuit}$\hspace{0.5mm}, Zhilin Wang$^\diamondsuit$\hspace{0.5mm} \\
  \bf{Longyue Wang$^{\heartsuit}$\hspace{0.5mm},
 Linyi Yang$^{\clubsuit}$\hspace{0.5mm},  
 Shuming Shi$^{\heartsuit}$\hspace{0.5mm}, 
 Yue Zhang$^{\clubsuit }$\footnotemark[2]}\hspace{0.2mm}\hspace{1.5mm} \\
$^\spadesuit$ Zhejiang University \ \ \ \quad$^\clubsuit$Westlake University \\ $^\vardiamondsuit$ The University of Hong Kong \ \ \ \quad$^\diamondsuit$Jilin University\ \ \ \quad$^\heartsuit$ Tencent AI lab 
 \\
 \texttt{yafuly@gmail.com} \quad \texttt{qtli@connect.hku.hk} \quad \\
 \texttt{nealcly.nlp@gmail.com} \quad \texttt{linzwcs@gmail.com} \\
 \quad\texttt{\{victoriabi,vinnylywang,shumingshi\}@tencent.com} \\
 \quad\texttt{\{yanglinyi,zhangyue\}@westlake.edu.cn}\\
}
\begin{document}
\maketitle
\renewcommand{\thefootnote}{\fnsymbol{footnote}}
\footnotetext[1]{\ Work was conducted during the internships of Yafu Li and Qintong Li at Tencent AI Lab.}
\footnotetext[2]{\ Corresponding authors.}
\begin{abstract}

Large language models (LLMs) have achieved human-level text generation,
emphasizing the need for effective AI-generated text detection to mitigate risks like the spread of fake news and plagiarism.
Existing research has been constrained by evaluating detection methods on specific domains or particular language models.
In practical scenarios, however, the detector faces texts from various domains or LLMs without knowing their sources.
To this end, we build a comprehensive testbed by gathering texts from diverse human writings and texts generated by different LLMs.
Empirical results show challenges in distinguishing machine-generated texts from human-authored ones across various scenarios, especially out-of-distribution. These challenges are due to the decreasing linguistic distinctions between the two sources.
Despite \textit{challenges}, the top-performing detector can identify 86.54\% out-of-domain texts generated by a new LLM, indicating the \textit{feasibility} for application scenarios.

\end{abstract}

\section{Introduction}

With constant advancements in artificial intelligence generated content (AIGC) technology \cite{diffusion,control-net,effidit,gpt3,gpt4},
texts generated by large language models (LLMs) \cite{gpt3,gpt4,llama,alpaca} have reached a level comparable to that of human peers, enabling the generation of remarkably fluent and meaningful responses to various user queries.

\begin{figure}
    \centering
\includegraphics[width=0.35\textwidth]{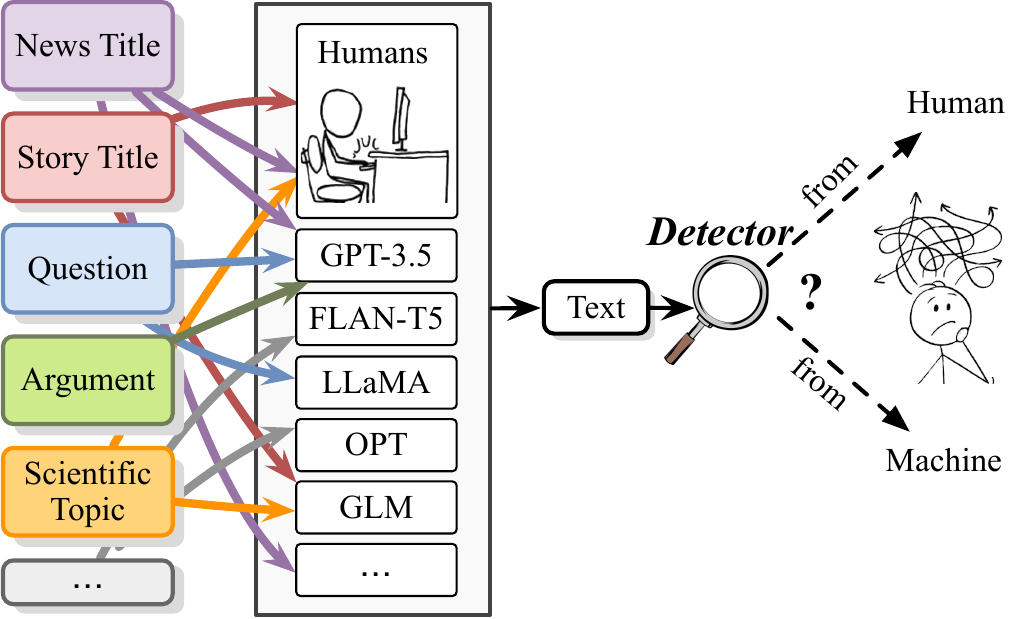}
\vspace{-0.1cm}
    \caption{Machine-generated text detection in the wild: the detector encounters texts from various human writings or fake texts generated by diverse LLMs.}
    \label{fig:intro}
\end{figure}

Advanced LLMs have become prevalent in enhancing human life and productivity. 
Nevertheless, they can also be employed for purposes such as manipulating public opinion, spreading fake news, and facilitating student plagiarism.
To this end, researchers have recently been putting efforts into differentiating between texts written by humans and those generated by machines \cite{jmpu_detect, chatgpt_detect, ft_water, detect-gpt}.
However, these findings are limited to testbeds of specific domains \cite{jmpu_detect} or deepfake texts from certain models \cite{chatgpt_detect}, or they assume the accessibility of the source LLMs \cite{ft_water, detect-gpt}.
Within a specific domain (e.g., BBC News), it can be easy to identify texts generated by a certain model (e.g., ChatGPT) from human writings~\cite{jmpu_detect,detect-gpt}.

In practice, however, a machine-generated text detector may encounter fake news from various LLMs without knowing their sources, as depicted in Figure~\ref{fig:intro}.
The detector can also face ChatGPT-generated student assignments across different tasks such as story generation, question answering, and scientific writing.
As the detector encounters increasingly diverse texts from both human-written and machine-generated sources, it has fewer surface patterns or linguistic differences to rely on.
In a more demanding scenario, the detector must identify texts from unfamiliar domains or those generated by unseen LLMs.
In this study, we try to address the following research questions: 
(1) Can existing detection methods effectively distinguish texts generated by diverse LLMs for various writing tasks in real-world scenarios?
(2) Are there inherent distinctions between human-written texts and machine-generated texts in an open-domain setting, irrespective of their topic or content?

To this end, we build a large-scale testbed, \textbf{MAGE}, for \textbf{MA}chine-\textbf{GE}nerated text detection, by collecting human-written texts from 7 distinct writing tasks (e.g., story generation, news writing and scientific writing) and generating corresponding machine-generated texts with 27 LLMs (e.g., ChatGPT, LLaMA, and Bloom) under 3 representative prompt types.
We categorize the data into 8 testbeds, each exhibiting progressively higher levels of ``wildness'' in terms of distributional variance and detection complexity.
Initially, we detect texts generated by a white-box LLM within a specific domain. 
Subsequently, we enhance the complexity by incorporating texts generated by additional LLMs across various writing tasks.
The most challenging testbed necessitates the detector's ability to identify out-of-domain texts generated by newly developed LLMs and perform detection against paraphrasing attacks.

We evaluate 4 commonly employed detection methods, encompassing both supervised and unsupervised approaches, on our proposed testbeds.
Empirical results indicate that all detection methods are effective in identifying machine-generated texts from a single domain or generated by a limited range of LLMs. 
However, as the diversity of domains and models increases, except for the PLM-based detector, all other methods experience significant performance deterioration. 
The challenge intensifies with out-of-distribution (OOD) testbeds, where even the best-performing detector misclassifies 61.95\% of human-written texts from unseen domains. 
The suboptimal OOD performance can be effectively mitigated by leveraging a mere 0.1\% of in-domain data, resulting in over 80\% recall for identifying out-of-domain texts generated by previously unencountered LLMs.
This demonstrates the feasibility of machine-generated text detection in real-world scenarios.

Finally, we investigate potential differences between human texts and machine generations that can be utilized for detection.
Statistical findings demonstrate that while significant linguistic differences exist within a particular domain, they gradually converge as more texts from diverse domains and language models are included.
Moreover, empirical results demonstrate that perplexity can serve as a fundamental feature for clustering the two sources of text. 
It is applicable to distinguishing between human and machine compositions in general, regardless of the text domain or the language model used for generation.
We release our resources at \url{https://github.com/yafuly/MAGE}.

\section{Related Work}

A line of work explores the linguistic patterns to achieve automatic machine-writing detection, which has gone through $n$-gram frequencies~\citep{badaskar2008identifying}, entropy~\citep{lavergne2008detecting,{gltr}}, perplexity~\citep{beresneva2016computer}, and negative curvature regions of the model's log probability~\citep{detect-gpt,fast-detect-gpt}. 
One limitation of these statistics-based methods is the white-box assumption that we can access the model prediction distributions, hindering wider applications on models behind APIs, such as ChatGPT. 
Another alternative paradigm is training neural-based detectors~\citep{bakhtin2019real,fagni2021tweepfake,uchendu2020authorship,openai_detector}. 
Some works~\citep{meral2009natural,krishna2023paraphrasing,ft_water,kirchenbauer2023watermark} explore the potential of watermarks in language models, 
making model-generated texts easier to detect.
~\citet{native} indicate that texts by non-native speakers are more likely to be incorrectly identified as AI-generated.
Our work does not assume language models are enhanced with watermarks, instead considering a more common detection setting where we do not know the sources of detected texts.

Current AI text detection has not achieved significant success, as evidenced by the successful exploits of paraphrasers that expose weaknesses in existing detectors~\citep{sadasivan2023can,krishna2023paraphrasing}, raising concerns about the robustness of current detection methods.
On the other hand, most of the detectors focus on specific domains, such as news~\citep{zellers2019defending,zhong2020neural} and reviews~\cite{chakraborty2023possibilities}, or specific models~\citep{jmpu_detect, rodriguez2022cross, detect-gpt}.
The transferability of detection capabilities to out-of-distribution scenarios, involving texts from unseen domains or models, remains uncertain and represents a crucial practical challenge.
To address this issue, we examine a scenario where texts from various domains generated by different language models are combined and extended to out-of-distribution settings with consideration for paraphrasing attacks.


\section{Dataset Construction}
\paragraph{Data Sourcing.}
We collect human-written texts from a set of benchmark datasets, which cover diverse writing tasks including:
(1) Opinion statement: 804 opinion statements from the /r/ChangeMyView (\textbf{CMV}) Reddit subcommunity \citep{CMV} and 1,000 reviews from \textbf{Yelp} dataset \citep{Yelp};
(2) News article writing: 1,000 news articles from \textbf{XSum} \citep{XSum}
and 777 news articles from TLDR\_news\footnote{https://huggingface.co/datasets/JulesBelveze/TLDR\_news}(\textbf{TLDR});
(3) Question answering: 1,000 answers from the \textbf{ELI5} dataset \cite{ELI5};
(4) Story generation: 1,000 prompted stories from the Reddit WritingPrompts (\textbf{WP}) dataset \citep{WP} and 1,000 stories from ROCStories Corpora (\textbf{ROC}) \citep{roc};
(5) Commonsense reasoning: 1,000 sentence sets for reasoning from \textbf{HellaSwag} \citep{hswag};
(6) Knowledge illustration: 1,000 Wikipedia paragraphs from \textbf{SQuAD} contexts \citep{SQuAD};
(7) Scientific writing: 1,000 abstracts of scientific articles from \textbf{SciXGen} \citep{scixgen}.


\paragraph{Model sets.}
We aim to adopt a wide spectrum of representative large language models (LLMs) to construct machine-generated texts. 
In particular, 
we consider 27 LLMs in this work: \textbf{OpenAI GPT} (text-davinci-002/text-davinci-003/gpt-turbo-3.5) \cite{gpt3}, \textbf{LLaMA} (6B/13B/30B/65B) \cite{llama}, \textbf{GLM-130B} \cite{glm}, \textbf{FLAN-T5} (small/base/large/xl/xxl) \cite{flant5}, \textbf{OPT} (125M/350M/1.3B/2.7B/6.7B/13B/30B/iml-1.3B/iml-30B) \cite{opt}, \textbf{BigScience} (T0-3B/T0-11B/BLOOM-7B1) \cite{t0,bloom} and \textbf{EleutherAI} (GPT-J-6B and GPT-NeoX-20B) \cite{gptj,gptneox}.

\paragraph{Prompts.}
To generate machine-generated text for each instance in the collected data, we use three types of prompts to feed the LLMs: (1) \textbf{continuation} prompts: ask LLMs to continue generation based on the previous 30 words of the original human-written text; (2) \textbf{topical} prompts: as LLMs to generate texts based on a topic (e.g., argument, news title, story topic, etc.) and (3) \textbf{specified} prompts: topical prompts with specified information about the text sources (e.g., BBC news, Reddit Post, etc.).
The topical and specified topical prompts are designed for OpenAI models, as they can respond to such prompts robustly. 
We present several prompt examples in Appendix \ref{app:prompts}.

In summary, for each human-written text, we generate a set of machine-generated texts using 27 LLMs with 3 different prompts. 
Data construction details and statistics are presented in Appendix \ref{app:data}.



\section{Detection Methods}
A detection system labels a text as either machine-generated or human-written, or outputs a probability distribution. 
In this work, we consider a set of commonly used detection methods. 
To showcase detection difficulty, we first consider naive baselines, i.e., \textbf{human detection} and \textbf{ask ChatGPT}, by asking human and query ChatGPT to identify the text source.
For supervised methods, we choose the \textbf{PLM-based classifier}, which is commonly used in text detection \cite{rodriguez2022cross,jmpu_detect}. 
We report the performance of Longformer~\cite{longformer} in the remainder of the paper, as it outperforms other commonly used PLMs, such as BERT~\cite{bert}, RoBERTa~\cite{roberta}, and GPT-2~\cite{gpt2}. Detailed comparisons can be found in Appendix~\ref{app:plm}.
\textbf{GLTR} \cite{gltr} is also included to represent methods that leverage model-based features.
In addition, we include \textbf{FastText} \cite{fasttext}, which uses linguistic statistics as features.
For unsupervised detection, we consider \textbf{DetectGPT} \cite{detect-gpt} to study the robustness of zero-shot detectors, which can also serve as a representative method that requires access to the text-generation LLM. 
Implementation details are shown in Appendix \ref{app:methods}.

\section{Experimental Setup}
\subsection{Testbed Settings} 
We consider each benchmark dataset as separate domains, such as CMV, XSum, SciXGen, etc. We group the LLMs into 7 sets based on their source: OpenAI GPT set, LLaMA set, GLM-130B set, FLAT-T5 set, OPT set, BigScience set, and EleutherAI set.
To investigate whether machine-generated text can be distinguished from human-written text, we categorize the collected data into \textbf{8 settings}. These settings are determined by the sources of training and evaluation data and increase in detection difficulty. The simplest setting involves detecting within-domain white-box detection while the most challenging setting involves detecting against paraphrasing attack.

We first consider \textbf{in-distribution settings}, where the detection method is evaluated on texts from seen domains and model sets, i.e., the training and test data are from the same data source.
\paragraph{Testbed 1: Fixed-domain \& Model-specific.}
Human-written texts come from a single domain and machine-generated texts are generated by a specific LLM (GPT-J-6B). A classifier is trained for each of the 10 domains, and the weighted average performance is reported. 
In this setting, we use only GPT-J-6B to generate fake texts instead of the entire model set from EleutherAI, 
aiming to simulate \textbf{white-box detection}, i.e., accessibility to the text-generating LLM, which is crucial for detection methods such as DetectGPT.

\paragraph{Testbed 2: Arbitrary-domains \& Model–specific.}
Human-written texts are obtained from combining all 10 domains, 
while machine-generated texts are produced by a single model set, creating 7 independent testbeds for each model set. 
We train 7 classifiers accordingly and report weighted average performance.

\paragraph{Testbed 3: Fixed-domain \& Arbitrary-models.}
Similarly, we include human-written texts from a single domain and obtain machine-generated using all model sets. 
In this way, we create 10 independent testbeds for each domain and train 10 classifiers accordingly.

\paragraph{Testbed 4: Arbitrary-domains \& Arbitrary-models.}
Human-written texts are from all domains with machine-generated texts generated using all model sets, which creates an integral testbed covering the full range of data.
We train a general classifier and report its performance. 

Furthermore, we consider four \textbf{out-of-distribution settings} where the detection model is tested on texts from unseen domains or unseen models.
\paragraph{Testbed 5: Unseen Models.}
This setting evaluates whether the classifier can detect texts from unseen models.
In this setting, texts generated by a specific model set are excluded from the training data. The classifier is then trained on the remaining texts and tested on the excluded ones. This process creates 7 testbeds for cross-validation.
We train 7 classifiers for each testbed and report their weighted average performance.

\paragraph{Testbed 6: Unseen Domains.}
This setting evaluates whether the classifier can detect texts from unseen domains.
In this setting, texts from a specific domain are excluded from the training data. The classifier is then trained on the remaining texts and tested on the excluded one. This process creates 10 testbeds for cross-validation.
We train 10 classifiers for each testbed and report weighted average performance.

\paragraph{Testbed 7: Unseen-domains \& Unseen-model.}
We go one step ``wilder'' by constructing an additional test set with texts from unseen domains generated by an unseen model, to test the detection ability in more practical scenarios. 
We consider four new datasets: CNN/DailyMail \cite{cnn}, DialogSum \cite{dialogsum}, PubMedQA \cite{pubmedqa} and IMDb \cite{imdb} to test the detection of machine-generated news, dialogues, scientific answers and movie reviews.
We sample 200 instances from each dataset and use a newly developed LLM, i.e., GPT-4 \cite{gpt4}, with specially designed prompts (Appendix \ref{app:prompts}) to create machine-generated texts.

\paragraph{Testbed 8: Paraphrasing Attack.}
\citet{sadasivan2023can} show that detection methods are vulnerable to being deceived by paraphrased target texts.
Based on the Unseen Domains \& Unseen Model test set, we paraphrase each sentence individually for both human-written and machine-generated texts, forming a more challenging test set. 
We treat paraphrases from both sources as machine-generated.
We adopt \texttt{gpt-3.5-turbo} as the paraphraser and consider all paraphrased texts as machine-generated.
\subsection{Evaluation Metrics}
We report \textbf{AUROC} (the area under the receiver operating characteristic
curve), which quantifies the classifier's potential of distinguishing between the positive and negative classes. 
An AUROC of 1.0 corresponds to a perfect classifier, whereas 0.5 represents random guessing.
Following \citet{rec1}, we also consider \textbf{AvgRec} (average recall), which is calculated by averaging the recall scores on human-written texts (HumanRec) and machine-generated texts (MachineRec) \footnote{Since our test sets are balanced, the precision score heavily relies on and can be reflected by the recall score. Therefore, we choose to report only the recall scores for a more intuitive evaluation.}. 
These recall scores help us assess the realistic detection performance.
For instance, black-box detection methods like human detection and ask ChatGPT cannot be evaluated using AUROC.
Furthermore, determining a decision boundary based on a reliable validation set is challenging in an open-domain detection setting.

\begin{table}[t]
\centering
\small
\begin{tabular}{lccc}
    \toprule
     \textbf{Detector} & \textbf{HumanRec}& \textbf{MachineRec} &\textbf{AvgRec}  \\
     \midrule
     ChatGPT & 96.98\% & 12.03\%   &  54.51\% \\
     Human &  61.02\% & 47.98\%   &  54.50\%   \\
\bottomrule
\end{tabular}
\vspace{-0.1cm}
    \caption{Detection performance of ChatGPT and humans.}
    \label{tab:human_gpt}
\end{table}

\begin{table}[t]
\centering
\small
\begin{tabular}{lccc}
    \toprule
     \textbf{Methods} & \textbf{Human}/\textbf{Machine} &\textbf{AvgRec} & \textbf{AUROC} \\ 
     \midrule

    FastText & 94.72\%/94.36\% & 94.54\% & 0.98 \\  
    GLTR & 90.96\%/83.94\% & 87.45\% & 0.94 \\  
    Longformer & 97.30\%/95.91\% & \textbf{96.60\%} & \textbf{0.99}  \\  
    DetectGPT & 91.68\%/81.06\% & 86.37\% & 0.92 \\
\bottomrule
\end{tabular}
\vspace{-0.3cm}
    \caption{(Testbed 1) White-box detection performance. ``Human/Machine'' denotes HumanRec and MachineRec, respectively.}
    \label{tab:white_box}
\end{table}

\begin{table*}[t]
\centering
\small
\begin{tabular}{cccccc}
\toprule
\multirow{2}{*}{\textbf{Settings}} &\multirow{2}{*}{\textbf{Methods}} & \multicolumn{4}{c}{\textbf{Metrics}} \\
\multicolumn{2}{c}{} & \textbf{HumanRec} & \textbf{MachineRec} & \textbf{AvgRec} & \textbf{AUROC} \\ 
\midrule 
\multicolumn{6}{c}{Testbed 2,3,4: In-distribution Detection}  \\

\midrule
& FastText \cite{fasttext}& 88.96\% & 77.08\% & 83.02\% & 0.89 \\  
Arbitrary-domains  & GLTR \cite{gltr} &75.61\% & 79.56\% & 77.58\% & 0.84 \\  
\&  Model–specific  & Longformer \cite{longformer} & 95.25\% & 96.94\% & \textbf{96.10\%} & \textbf{0.99} \\  
& DetectGPT{$^\star$} \cite{detect-gpt} & 48.67\% & 75.95\% & 62.31\% & 0.60 \\ 
%
\midrule
& FastText \cite{fasttext} & 89.43\% & 73.91\% & 81.67\% & 0.89 \\  
Fixed-domain & GLTR \cite{gltr} & 37.25\% & 88.90\% & 63.08\% & 0.80 \\  
\&  Arbitrary-models & Longformer \cite{longformer} & 89.78\% & 97.24\% & \textbf{93.51\%} & \textbf{0.99}\\  
& DetectGPT{$^\star$} \cite{detect-gpt} & 86.92\% & 34.05\% & 60.48\% & 0.57 \\ 
%
\midrule
& FastText \cite{fasttext} & 86.34\% & 71.26\% & 78.80\% & 0.83 \\  
Arbitrary-domains & GLTR \cite{gltr} & 12.42\% & 98.42\% & 55.42\% & 0.74 \\  
\& Arbitrary-models & Longformer \cite{longformer} & 82.80\% & 98.27\% & \textbf{90.53\%} & \textbf{0.99} \\ 
 
& DetectGPT{$^\star$} \cite{detect-gpt} & 86.92\% & 34.05\% & 60.48\% & 0.57 \\
%
\midrule
\multicolumn{6}{c}{Testbed 5,6: Out-of-distribution Detection} \\
\midrule
\multirow{4}{*}{Unseen Models} 
& FastText \cite{fasttext} & 83.12\% & 54.09\% & 68.61\% & 0.74 \\  
& GLTR \cite{gltr} & 25.77\% & 89.21\% & 57.49\% & 0.65 \\  
& Longformer \cite{longformer} & 83.31\% & 89.90\% & \textbf{86.61\%} & \textbf{0.95} \\  
& DetectGPT{$^\star$} \cite{detect-gpt} & 48.67\% & 75.95\% & 62.31\% & 0.60 \\ 
\midrule
\multirow{4}{*}{Unseen Domains} 
& FastText \cite{fasttext} & 54.29\% & 72.79\% & 63.54\% & 0.72 \\  
& GLTR \cite{gltr} & 15.84\% & 97.12\% & 56.48\% & 0.72 \\  
& Longformer \cite{longformer} & 38.05\% & 98.75\% & \textbf{68.40\%} & \textbf{0.93} \\  
& DetectGPT{$^\star$} \cite{detect-gpt} & 86.92\% & 34.05\% & 60.48\% & 0.57 \\
%
%
\bottomrule
\end{tabular}
\vspace{-0.2cm}
\caption{(Testbed 2-6) Detection performance of different detection methods. The out-of-distribution settings examine the detection capability on texts from unseen domains or machine-generated texts generated by new LLMs. $\star$ denotes the unsupervised detection method. }
\label{tab:main}
\end{table*}

\section{Results}

\subsection{Naive Baselines}
\label{sec:human_detect}
Table \ref{tab:human_gpt} shows that both ChatGPT and human annotators fail to distinguish machine-generated texts from human-written ones.
The AvgRec is only slightly better than random guessing, suggesting that machine-generated texts have achieved a level (e.g., fluency and coherence) comparable to those of humans.
We then explore whether there exist underlying differences that can be captured by automatic detection methods.

\subsection{In-domain Detection}
The results of in-domain detection are shown in Table~\ref{tab:white_box} and the upper part of Table~\ref{tab:main}.

\paragraph{White-box Detection.} 
From Table~\ref{tab:white_box}, we can observe that all detection methods obtain solid performance when the texts are from a specific domain and a specific LLM (GPT-J-6B) (i.e., \textit{Fixed-domain \& Model-specific}).
Typically, DetectGPT performs well in identifying machine-generated texts when the scoring model matches the one used to generate the fake texts, i.e., accessibility to the generation LLM in the white-box setting.

\paragraph{PLM-based Detectors demonstrate robustness to texts from various sources.}
As shown in Table~\ref{tab:main}, the detection performance (AvgRec and AUROC) decreases as the detector encounters broader data sources, i.e., texts from various domains or various LLMs.
For example, GLTR's AUROC drops from 0.94 to 0.80 and DetectGPT's drops from 0.92 to 0.57 when encountering texts from multiple models (\textit{Arbitrary-models}). 
The severe performance drop of DetectGPT is attributed to its reliance on accessibility to the generation LLMs~\cite{detect-gpt}.
On the other hand, FastText faces significant challenges in detecting texts from various domains (\textit{Arbitrary-domains}), despite its robustness on texts sourced by different language models.
Among all detection methods, the Longformer detector consistently outperforms others in terms of AUROC and AvgRec.
Despite the minor performance degradation, Longformer surpasses other detectors by a considerable margin in the \textit{Arbitrary-domains \& Arbitrary-models} setting, where the detector encounters diverse texts from various domains and language models.


\subsection{Out-of-domain Detection}
We further investigate whether the detection model can identify machine-generated texts in out-of-distribution settings, i.e., detect texts from unseen domains or generated by new LLMs.
The results are presented in the lower part of Table \ref{tab:main}.
Empirical results indicate that, except for the Longformer detector, all other detectors perform poorly in identifying texts generated by unseen models.
Furthermore, none of the detectors effectively classify texts from novel domains.

\paragraph{Unseen Models.}

\begin{figure}[t!]
\centering
\includegraphics[width=0.8\linewidth]{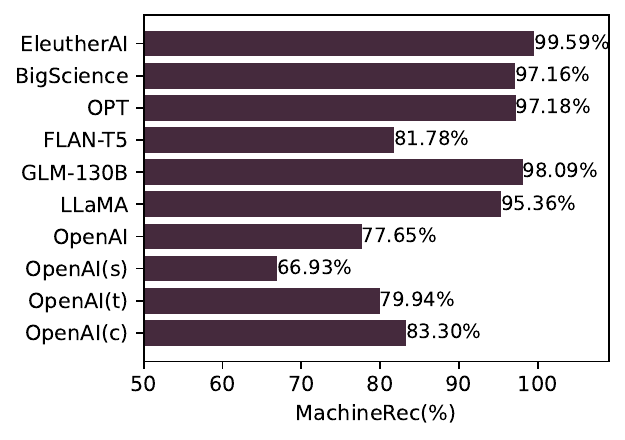}
\vspace{-0.1cm}
    \caption{Out-of-distribution detection performance on machine-generated texts generated by \textit{unseen models}. OpenAI(c), OpenAI(t) and OpenAI(s) corresponds to texts generated by OpenAI models using continuation, topical and specified prompts, respectively.}
    \label{fig:model_set}
\end{figure}


Among all methods, the Longformer detector is the only one that performs well (with an AUROC of 0.95 and AvgRec of 86.61\%) when detecting texts from unseen LLMs.
The performance of FastText further degrades, with AUROC dropping from 0.83 to 0.74.
GLTR faces a significant challenge when it comes to unseen models. 
Its AUROC of 0.65 suggests that it struggles to differentiate between different text sources.
The detection performance (Longformer) on each unseen model set is shown in Figure \ref{fig:model_set}.
The Longformer classifier has the most difficulty distinguishing texts generated by the OpenAI and FLAN-T5 models from human-written ones.
By comparison, the detector can identify most of the machine-generated texts from other models, even if it has not encountered any of them during training.
On the other hand, the difficulty of detection is influenced by the prompt types used for model generation. Texts generated from specific prompts (OpenAI(s)) are harder to distinguish than continuation prompts (OpenAI(c)) and topical prompts (OpenAI(t)). This can be because they follow a detailed prompt condition, making them more similar to human-written texts.



\paragraph{Unseen Domains.}
\begin{figure}[t]
\vspace{-0.cm}
\centering
\includegraphics[width=0.8\linewidth]{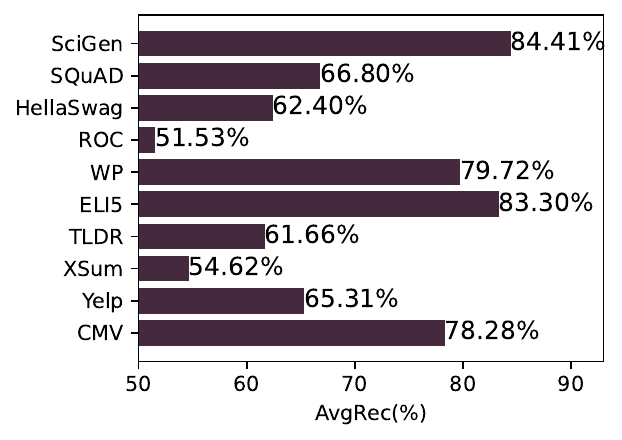}
\vspace{-0.1cm}
    \caption{Out-of-distribution detection performance (AvgRec) on texts from \textit{unseen domains}.}
    \label{fig:domain_set}
\end{figure}
Detecting texts from \textit{unseen domains} presents a heightened challenge for classifiers. Notably, even the top-performing model, Longformer, experiences a substantial decline in AvgRec, dropping from 90.53\% to 68.40\%.
Typically, Longformer tends to classify human-written texts from unfamiliar domains as machine-generated, 
which results in a low HumanRec score but an almost perfect MachineRec. 
We present detection performance (Longformer) on each unseen domain in Figure \ref{fig:domain_set}.
The top three text domains most likely to be misclassified as machine-generated are ROC, XSum, and TLDR datasets. 
This could be attributed to their low average perplexity scores which confuse PLM-based detectors (discussed in Section \ref{sec:ppl_bias}).


\begin{figure}[t!]
    \centering
    \begin{subfigure}[t]{0.48\linewidth}
        \centering
        \includegraphics[width=\linewidth]{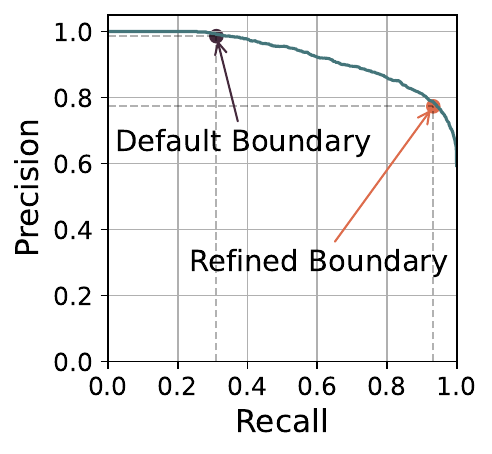}
        \caption{Precision-Recall curve of the Longformer detector on the unseen domain (Yelp). A refined decision boundary obtains a better trade-off between precision and recall.}
        \label{fig:decision_merge1}
    \end{subfigure}
    \hfill
    \begin{subfigure}[t]{0.48\linewidth}
        \centering
        \includegraphics[width=\linewidth]{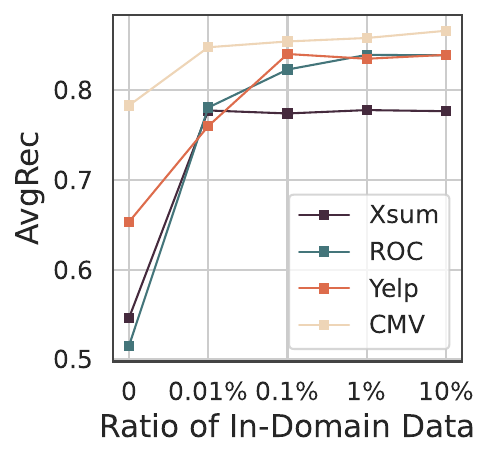}
        \caption{Detection performance in the "Unseen Domains" setting (Xsum, ROC, Yelp and CMV) with decision boundary adjusted based on different ratios of in-domain data.}
        \label{fig:decision_merge2}
    \end{subfigure}
    \caption{Decision boundary adjustment.}
    \label{fig:decision_merge}
\end{figure}



\begin{table}[t]
\centering
\small
\begin{tabular}{c|c|c}
    \toprule
     \textbf{Metrics} & \textbf{Unseen Models}& \textbf{Unseen Domains}  \\
     \midrule
      HumanRec & 86.09\%   &  82.88\% \\
      MachineRec & 89.15\%   &  80.50\% \\
      AvgRec &\textbf{ 87.62\%(+1.01\%)}   &  \textbf{81.78\%(+13.38\%)} \\
\bottomrule
\end{tabular}
\vspace{-0.2cm}
    \caption{Detection performance (Longformer) on out-of-distribution testbeds with decision threshold adjusted based on 0.1\% of the in-distribution data.}
    \label{tab:re_bound}
\end{table}

\paragraph{Boundary Adjustment.}
Despite the low AvgRec in the \textit{Unseen Domains} setting, Longformer achieves a high AUROC score (0.93). This suggests that the model can distinguish between the two classes but struggles with selecting an appropriate decision boundary, as shown in Figure \ref{fig:decision_merge1}.
To address this issue, we utilize a portion of the in-domain data from the training set to adjust the decision boundary.
We compute an average decision boundary across 10 classifiers (in the \textit{Unseen Domains} setting) and apply it universally across all domains.
As depicted in Figure \ref{fig:decision_merge2}, refining the decision boundary with only 0.1\% of in-domain data (e.g., 4 instances for CMV) significantly enhances detection performance.
Table \ref{tab:re_bound} demonstrates that adjusting the decision boundary (using 0.1\% of in-domain data) notably improves detection accuracy for both out-of-distribution settings.

\paragraph{Unseen Domains \& Unseen Model}
We validate the detection ability of Longformer, the best-performing detector, on the \textit{Unseen Domains \& Unseen Model} testbed. The results are presented in Table \ref{tab:wilder_testbeds}.
The Longformer detector trained using our dataset achieves a high performance (0.94 AUROC) in detecting texts generated by GPT-4, even when sourced from newly added datasets and generated by a new LLM.
After refining the boundary, the detector demonstrates balanced accuracy in detecting both text sources, resulting in an AvgRec of 86.54\%. This showcases its feasibility for deployment in real-world scenarios.

\paragraph{Paraphrasing Attack}
However, similar to other methods~\cite{krishna2023paraphrasing}, the Longformer detector also shows vulnerability to paraphrasing attacks, as shown in Table \ref{tab:wilder_testbeds}.
The AUROC drops from 0.94 to 0.75 when the detector encounters additional paraphrased texts, which can be attributed to the shifted perplexity distribution of paraphrased texts (Section \ref{sec:ppl_bias}).

\begin{table}[t]
\centering
\small
\begin{tabular}{cccc}
    \toprule
     \textbf{HumanRec} & \textbf{MachineRec}& \textbf{AvgRec}  & \textbf{AUROC} \\
    \midrule
    \multicolumn{4}{c}{Testbed 7: Unseen Domains \& Unseen Model }\\
    \midrule
     52.50\% & 99.14\% & 75.82\% & 0.94 \\
     88.78\dag & 84.12\%\dag & 86.54\%\dag & 0.94 \\
     \midrule
     \multicolumn{4}{c}{Testbed 8: Paraphrasing Attack}\\
     \midrule
     52.16\% & 81.73\% & 66.94\% & 0.75 \\
     88.78\%\dag & 37.05\%\dag & 62.92\%\dag & 0.75 \\

\bottomrule
\end{tabular}
\vspace{-0.1cm}
    \caption{(Testbed 7-8) Detection performance of Longformer detector on the two challenging test sets. \dag  denotes the refined decision boundary.
    Appendix~\ref{app:ood_para_perform} includes the performance of other detection methods.}
    \label{tab:wilder_testbeds}
\end{table}

\section{Analysis}

\begin{figure}[t!]
\centering
\includegraphics[width=0.99\linewidth]{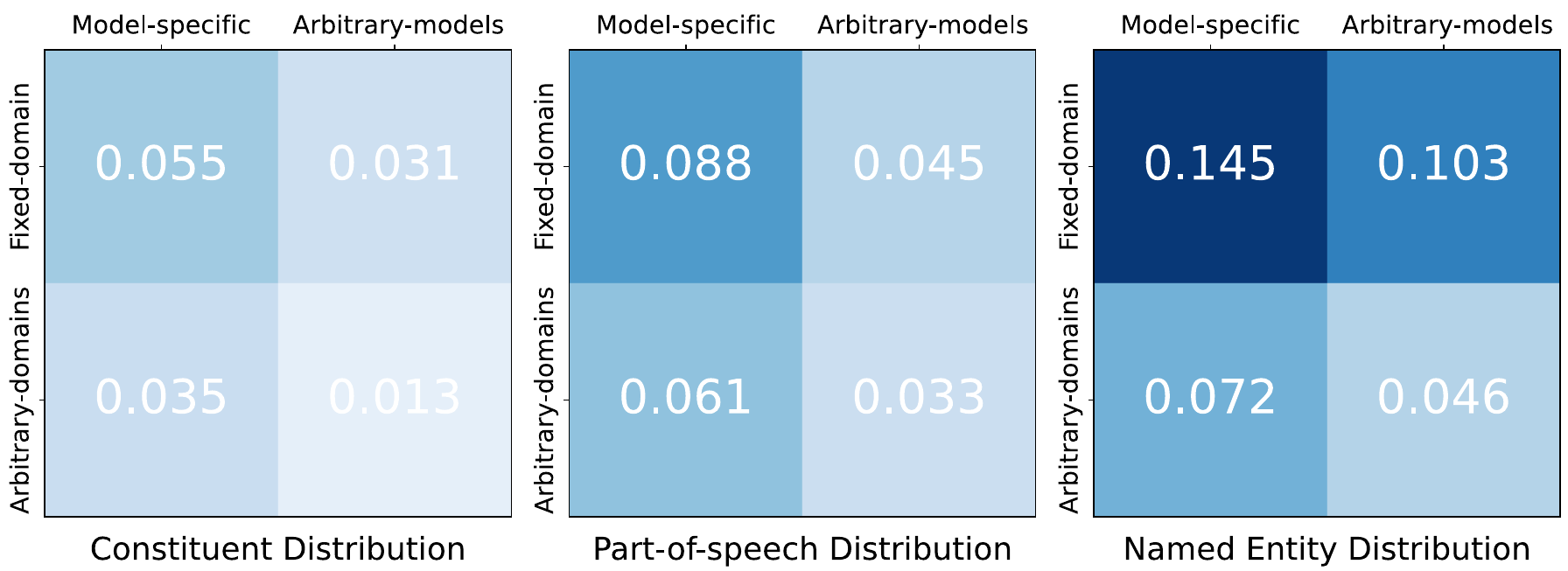}
    \vspace{-0.5cm}
    \caption{Linguistic difference (Jensen-Shannon distance) between human-written texts and machine-generated texts in 4 in-distribution settings (darker colors indicate larger differences).}
    \label{fig:js}
\end{figure}

\begin{figure}[t!]
\centering
\includegraphics[width=0.9\linewidth]{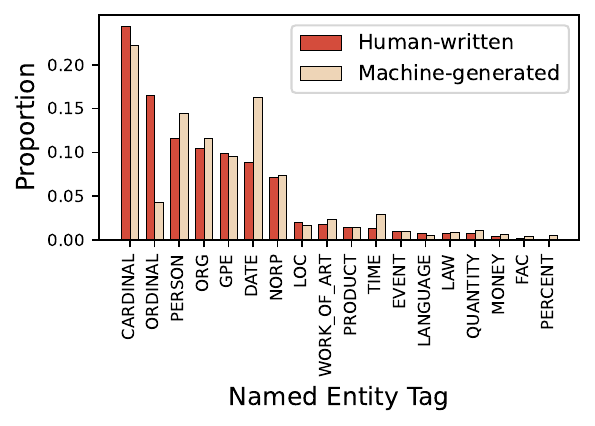}
    \vspace{-0.1cm}
    \caption{Linguistic difference (Named Entity Distributions) of the \textit{Fixed-domain \& Model-specific} setting.}
    \label{fig:js_ner}
\end{figure}

\subsection{Convergence of Human and Machine Compositions}
We explore to find potential differentiability through a comparison of linguistic patterns in human-written and machine-generated compositions.
To accomplish this, we employ Stanza~\cite{stanza} to extract the distribution of various linguistic patterns such as named entities, part-of-speech tags, and constituents.
Next, we calculate the Jensen-Shannon distance to quantify the disparity between the probability distributions obtained from both text sources (human-written and machine-generated).

Figure \ref{fig:js} demonstrates that including texts from diverse domains and LLMs reduces the linguistic dissimilarity between the two text sources.
This makes it more challenging for a detector to distinguish them, which aligns with the increasing difficulty of detection in the four in-distribution settings.
Once an adequate amount of texts from various domains and LLMs are collected, there is no significant statistical distinction between the two text sources (see Figure \ref{fig:ling} in Appendix \ref{app:stat}).
In contrast, when dealing with texts from a specific domain or an LLM (Fixed-domain \& Model-Specific), noticeable differences exist. For example, entity tags like "ORDINAL" and "DATE" can serve as detection shortcuts, as shown Figure~\ref{fig:js_ner}.
Comparing the sentiment polarity and grammatical formality of the two text sources (Appendix ~\ref{app:stat}) also demonstrates convergence between human-written and machine-generated texts.


\begin{figure}[t!]
\centering
\includegraphics[width=0.6\linewidth]{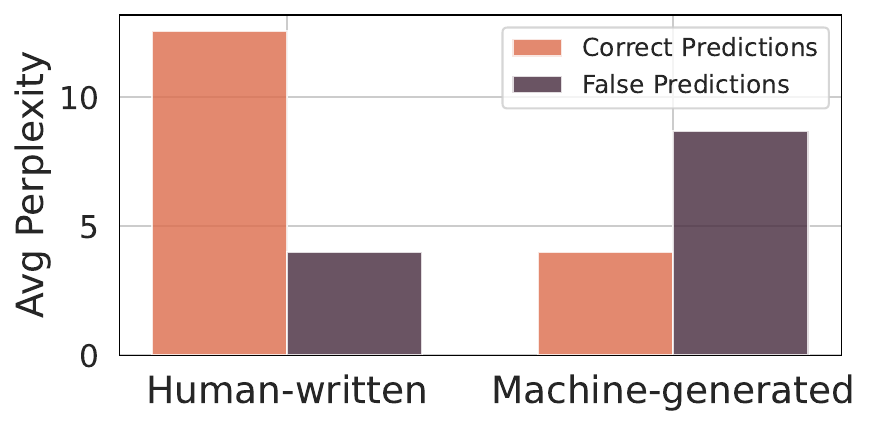}
    \vspace{-0.3cm}
    \caption{Comparison of the average perplexity of texts which the Longformer detector predicts correctly and incorrectly.}
    \label{fig:ppl1}
\end{figure}

\begin{figure}[t!]
\centering
\includegraphics[width=0.8\linewidth]{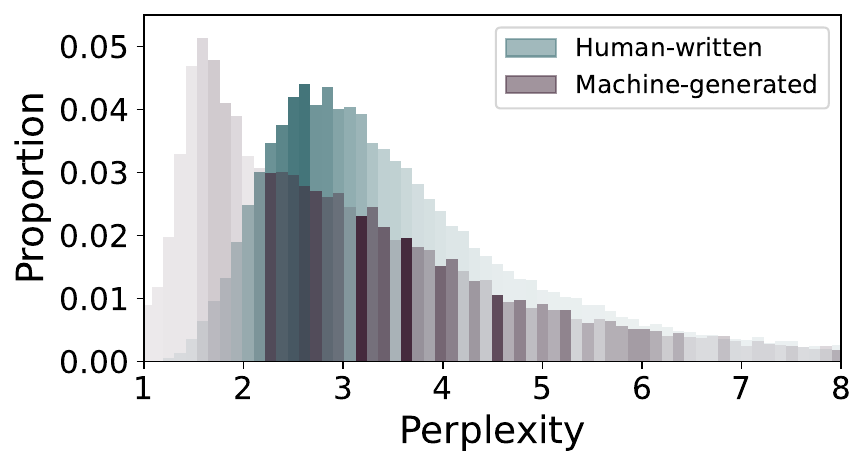}
    \vspace{-0.3cm}
    \caption{Perplexity distribution: A darker colour indicates a larger proportion of incorrect predictions in the perplexity bucket.}
    \label{fig:ppl2}
\end{figure}

\begin{figure}[t!]
\setlength{\belowcaptionskip}{-0.3cm}
\centering
\includegraphics[width=0.8\linewidth]{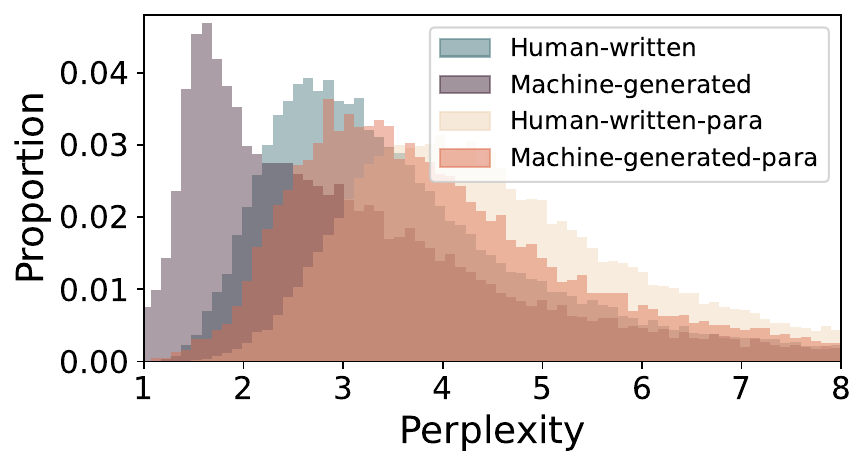}
    \vspace{-0.3cm}
    \caption{Perplexity distribution of human-written texts, machine-generated texts and their corresponding paraphrased texts.}
    \label{fig:ppl2_para}
\end{figure}

\subsection{Double-edged Sword of Perplexity Bias}
\label{sec:ppl_bias}
In this section, we explore to find the general distinction which is not influenced by text domain or generation LLMs. 
Prior work on unsupervised detection~\cite{detect-gpt, fast-detect-gpt} leverages the property that model generations reside in local minima of perplexity.
We discover that such property also acts as a fundamental feature for PLM-based methods to effectively differentiate machine generations.


Specifically, we use an \textbf{untuned} Longformer to obtain perplexity score \cite{ppl_mlm} for test set texts in the \textit{Unseen Domains} setting.
Figure~\ref{fig:ppl1} illustrates how prior knowledge in PLMs, as measured by perplexity, aids in clustering two text sources into distinct peaks. 
The average perplexity score of machine-generated texts is notably lower than that of human writings, establishing an implicit pattern to distinguish them. 

However, perplexity bias can hinder robust detection.
PLM-based detectors also exhibit overconfidence in text perplexity, classifying low-perplexity texts as machine-generated and high-perplexity texts as human-generated.
We categorize the texts based on prediction correctness. 
As shown in Figure \ref{fig:ppl1}, misidentified human-written texts by the Longformer detector have significantly lower average perplexity compared to correctly predicted ones, but are similar to correctly predicted machine-generated texts.
In contrast, the average perplexity of incorrectly predicted machine-generated texts is higher than that of correctly predicted ones.
Figure \ref{fig:ppl2} presents a more intuitive visualization: false predictions of human-written texts (darker green bars) are concentrated in the lower perplexity region, while false predictions of machine-generated texts (darker khaki bars) are spread across the higher perplexity region. 
Paraphrasing attacks, illustrated in Figure~\ref{fig:ppl2_para}, cause the peak of human-written texts to be positioned between that of machine-generated texts (machine-generated, machine-generated-para, and human-written-para), leading to significant confusion for the Longformer detector.

\section{Conclusion}
We proposed a comprehensive testbed for machine-generated text detection, by gathering texts from various writing tasks and machine-generated texts generated by different LLMs.
Empirical results on commonly used detection methods demonstrated the challenge of AI-generated text detection.
Out-of-distribution posed a greater challenge for detectors to be employed in application scenarios.
With the boundary refined, the best-performing detector on our testbeds (i.e., Longformer detector) achieved 86.54\% AvgRec on out-of-domain texts generated by a new LLM, i.e., GPT4.
By studying differences between human and machine compositions, we find that perplexity can serve as a fundamental feature for classification regardless of text domain or generation LLM.
To the best of our knowledge, this is the first study to investigate the challenges and feasibility of AI-generated text detection in a "wild" testbed.

\section*{Limitations}
Although we are the first to propose a comprehensive testbed for AI-generated text detection and validate the detection effectiveness on frontier test sets, there are two major limitations:
(1) We strive to include a wide variety of LLMs in our dataset. However, new LLMs such as Alpaca \cite{alpaca} and Vicuna \cite{vicuna2023} continue to emerge and may not be currently included. 
Nevertheless, our dataset aims to serve as a testbed to select the best-performing detectors, which encounter sufficiently diverse machine-generated texts and can deal with texts from newly-developed LLMs in future.
(2) We adopt benchmark datasets as text sources, which can be used as the training data for LLM pretraining. 
The detection capability may vary on new online texts that were not included in the LLMs' pretraining data.
In the future, we plan to gather new online texts that have not been previously seen by LLMs to study such variation.

\section*{Ethics Statement}
We honor the Code of Ethics. 
No private data or non-public information is used in this work.
For human annotation (Section~\ref{sec:human_detect}), we recruited our annotators from the linguistics departments of local universities through public advertisement with a specified pay rate. 
All of our annotators are senior undergraduate students or graduate students in linguistic majors who took this annotation as a part-time job. 
We pay them 60 CNY an hour. The local minimum salary in the year 2023 is 25.3 CNY per hour for part-time jobs.
The annotation does not involve any personally sensitive information. 

\section*{Acknowledgement}
We would like to thank all reviewers for their insightful comments and suggestions to help improve the paper. 
This work has emanated from research conducted with the financial support of the National Natural Science Foundation of China Key Program under Grant Number 62336006.


\bibliographystyle{acl_natbib}
\bibliography{emnlp2023}

\clearpage

\appendix

\begin{figure*}[t!]
\small
\centering
\includegraphics[width=0.9\linewidth]{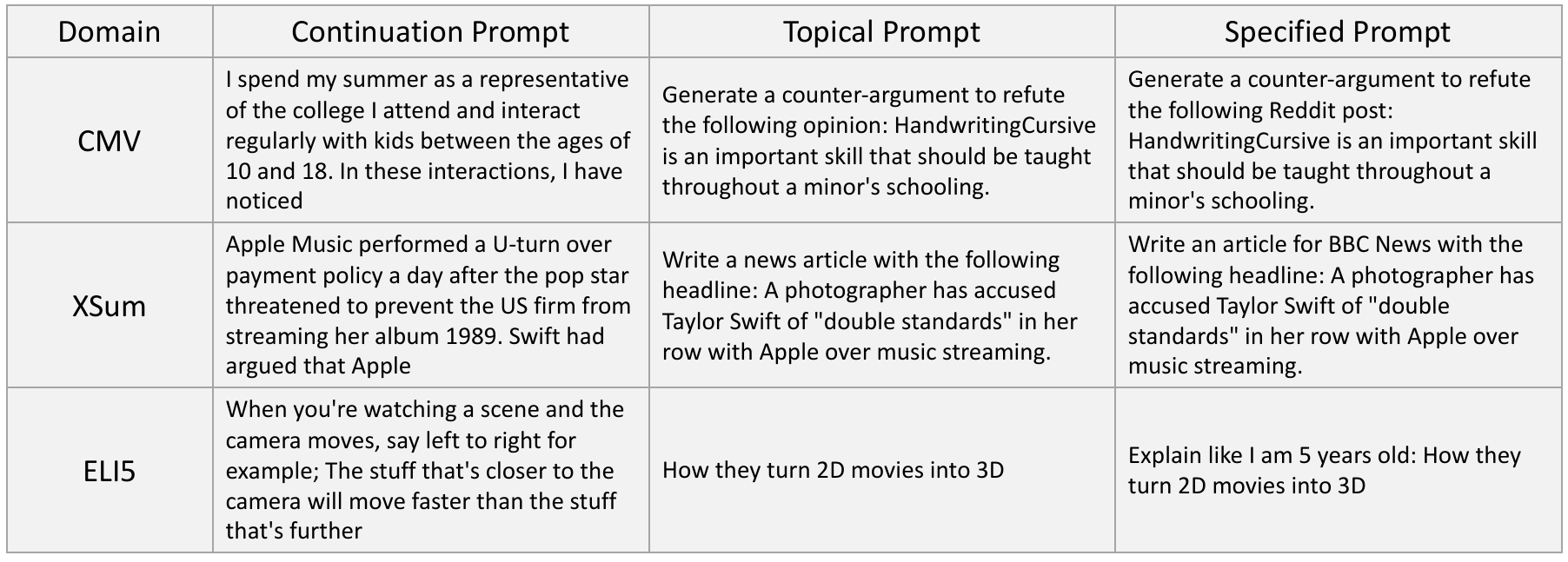}
\caption{
\label{fig:prompt}
Examples of three prompt types.}
\end{figure*}

\begin{figure*}[t!]
\small
\centering
\includegraphics[width=0.9\linewidth]{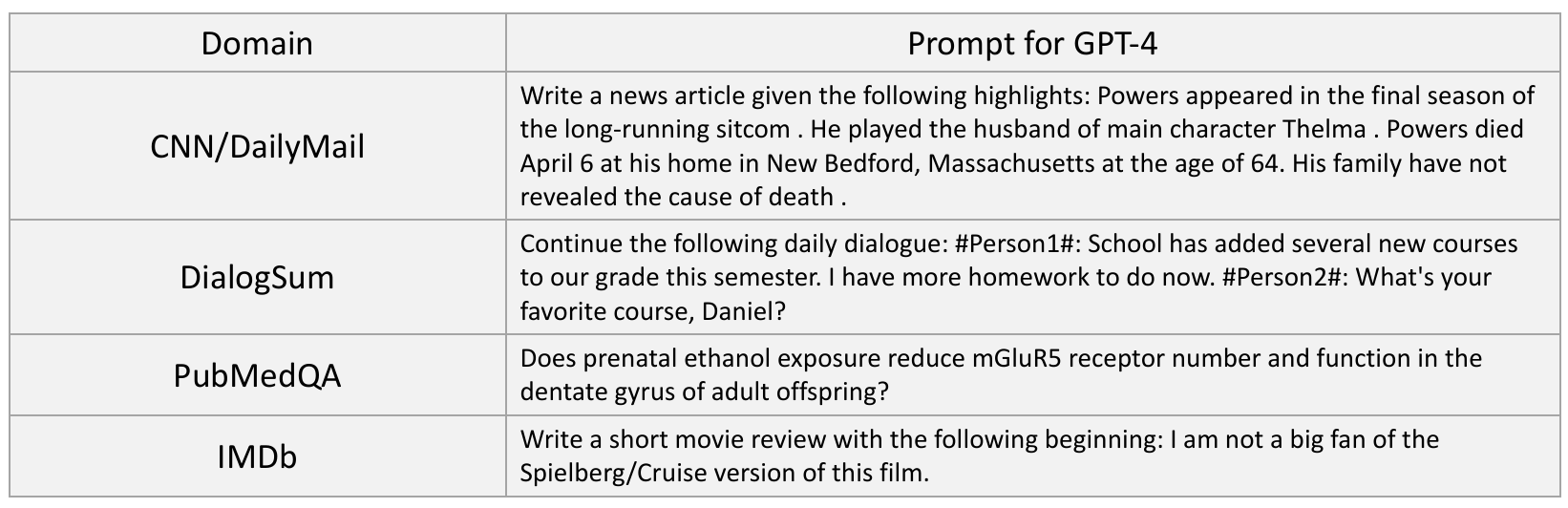}
\caption{
\label{fig:prompt2}
Examples of prompts for building the frontier test sets.}
\end{figure*}

\section{Prompt Design}
\label{app:prompts}
Figure \ref{fig:prompt} present prompt cases in three domains (CMV, XSum and ELI5) to showcase different prompt types (i.e., continuation prompts, topical prompts and specified prompts).
The prompts used for building GPT-4 test sets are presented in Figure \ref{fig:prompt2}.

\section{Dataset Construction}
\label{app:data}
We show an example of Yelp dataset to give an intuitive illustration of dataset construction: 
We randomly sample 1,000 human-written texts from the Yelp dataset and use 27 LLMs to generate corresponding machine-generated texts.
After data preprocessing and filtering, we obtained a total of 26,235 machine-generated texts and 1,000 human-written texts.
To mitigate data imbalance between the text sources (human-written v.s. machine-generated), we additionally collect data from the Yelp dataset and obtain a total of  37,706 human-written texts after filtering.
The additional data is used to compensate validation and test sets first for more accurate evaluation.
We discuss the effects of data balance for training in Appendx~\ref{app:balance}.

By default, machine-generated texts are generated using continuation prompts.
For datasets which provide topics or titles, we also consider topical and specified prompts. 
The latter two prompt types are only used for the OpenAI GPT model set, since we empirically find they perform robust generation to various prompts.
For example, for the 1,000 human-written texts in the Xsum dataset, we have 33,000 (27,000+3*2*1000) machine-generated texts and finally obtain 32,930 texts after filtering.

\begin{figure*}[t!]
\small
\centering
\includegraphics[width=0.8\linewidth]{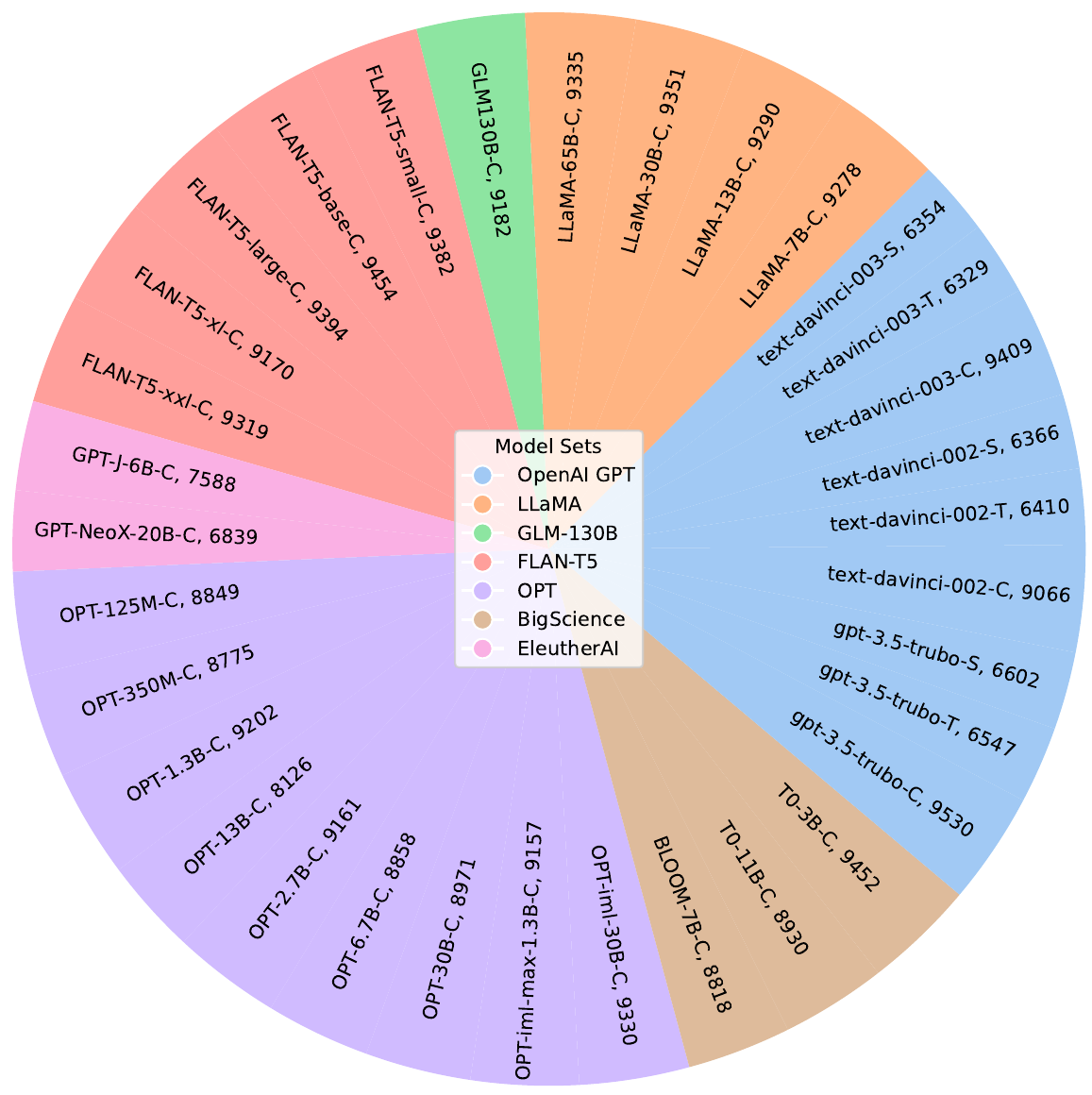}
\caption{
\label{fig:model_dtb}
Distribution of machine-generated instances by model: For example, "FLAN-T5-small-C, 9382" indicates that the model "FLAN-T5-small" generated 9382 texts using \textbf{c}ontinuation prompts. The letters \textbf{C}, \textbf{T} and \textbf{S} represent the types of prompts used: "\textbf{c}ontinuation" "\textbf{t}opical" and "\textbf{s}pecified", respectively.}
\end{figure*}

We conduct preprocessing to reduce the effects beyond text contents, such as punctuation normalization and line-break removal, etc.
We also filter out texts that are too long or too short. 
We divide the texts into three splits, i.e., train/validation/test, with an 80\%/10\%/10\% partition.
The data statistics are shown in Table \ref{tab:data_stat}.
The distribution of machine-generated texts by model is presented in Figure~\ref{fig:model_dtb}.

\begin{table*}[t!]
\small
    \centering
    \begin{tabular}{cccccc}

    \toprule
         \textbf{Dataset} & \textbf{CMV} & \textbf{Yelp} & \textbf{XSum} & \textbf{TLDR} & \textbf{ELI5}\\
         \midrule
         Train & 4,461/21,130 & 32,321/21,048 & 4,729/26,372 & 2,832/20,490  & 17,529/26,272  \\ 
         Valid & 2,549/2,616 & 2,700/2,630 & 3,298/3,297 & 2,540/2,520  & 3,300/3,283  \\
         Test & 2,431/2,531 & 2,685/2,557 & 3,288/3,261 & 2,536/2,451  &  3,193/3,215  \\
         \midrule
        \textbf{WP} & \textbf{ROC} & \textbf{HellaSwag} & \textbf{SQuAD} &\textbf{ SciXGen} & \textbf{all} \\
         6,768/26,339 & 3,287/26,289 & 3,129/25,584 & 15,905/21,489 & 4,644/21,541 & 95,596/236,554 \\
           3,296/3,288 & 3,286/3,288 & 3,291/3,190 & 2,536/2,690 & 2,671/2,670  & 29,467/29,462 \\
            3,243/3,192 & 3,275/3,207 & 3,292/3,078 & 2,509/2,535 & 2,563/2,338 & 29,015/28,365 \\
     \bottomrule
     
    \end{tabular}
    \caption{Number of instances for each dataset. The number of human-written texts and that of machine-generated texts are separated by "/".} 
    \label{tab:data_stat}
\end{table*}

\begin{table*}[h]
\small
\centering
\begin{tabular}{lccccccccccc}
\toprule
& CMV & Yelp & XSum & TLDR & ELI5 & WP & ROC & HellaSwag & SQuAD & SciXGen & all \\
\midrule
\# human & 80 & 100 & 100 & 77 & 100 & 100 & 100 & 100 & 100 & 99 & 1912 \\
\# machine & 80 & 100 & 100 & 77 & 100 & 100 & 100 & 100 & 100 & 99 & 1912 \\
\bottomrule
\end{tabular}
\caption{Number of human-written and machine-generated texts of the sampled testset for naive baselines.}
\label{tab:data_naive}
\end{table*}

\section{Method Implementation}
\label{app:methods}
\paragraph{Human annotation \& Ask-ChatGPT.}
We create a test subset from the whole testset, by pairing one machine-generated text with each human-generated one through random sampling.
To create the test set for the naive baselines, we randomly select 10\% of the human-written texts from the test set used in the "Arbitrary-domains \& Arbitrary-models" setting. 
Data statistics of the test set is shown in Table~\ref{tab:data_naive}.
We also randomly sample an equal number of machine-generated texts. 
We hire 3 expert annotators to conduct independent annotation and average their performance.

\paragraph{Longformer.} Across all datasets, we used the Adam optimizer \cite{adam} with
a learning rate of 0.005 and set the dropout rate at 0.1.
All models are finetuned for 5 epochs on 8 V100 GPUs.
We select the best-performing model based on validation classification accuracy.

\paragraph{FastText.}
We experiment with different combinations of word n-gram features and character n-gram features.
Based on validation results, we choose only word bi-grams as text features.
We train all models for 100 epochs and leave other settings as default.

\paragraph{GLTR.}
GLTR uses a language model to gather features, i.e., the number of tokens in the Top-10, Top-100, and Top-1000 ranks, 
which are fed into a logistic regression model to classify texts.
Following \citet{jmpu_detect}, we use GPT-2-XL \cite{gpt2} as the language model and use scikit-learn \cite{scikit-learn} to train regression models.
We conduct a grid search on optimization algorithm ('lbfgs', ``liblinear'', ``newton-cg'', ``newton-cholesky'', ``sag'', and ``saga''),  the norm of the penalty (``l1'', ``l2'' and ``elasticnet'') and regularization strength (0.001, 0.01, 0.1, 1, 10, and 100) and choose the best-performing model under cross-validation.

\paragraph{DetectGPT.}
We follow the best-performing setting \cite{detect-gpt}, using T5-3B \cite{2020t5} as the mask infilling model, with the mask rate set as 15\%, the masked span length as 2, and the number of perturbations as 100.
We use GPT-J-6B \cite{gptj} as the scoring model.
We manually set the decision boundary based on the validation set.

\section{Randomness}
\label{app:randomness}
We conduct experiments to testify the stability of our testbeds. Specifically, we investigate the effects of randomness under the \textit{Arbitrary-domains and Arbitrary-models} setting by (1) splitting the testbeds (train, validation and test) with 5 different seeds and training 5 Longformer detectors on each split; and (2) training 5 Longformer detectors with different running seeds on one of the splits.
The results in Table \ref{tab:random} show that our testbeds are robust to randomness, with a small standard deviation.

\begin{table*}[t!]
\centering
\small
\begin{tabular}{ccccc}
    \toprule
    \textbf{Randomness} &  \textbf{HumanRec}& \textbf{MachineRec} &\textbf{AvgRec} & \textbf{AUROC} \\
     \midrule
     Data Split & 83.00\%$\pm$2.82\% & 97.74\%$\pm$0.34\% & 90.37\%$\pm$1.29\%   &  0.99$\pm$0.0010 \\
     \midrule
     Training (Longformer) & 82.81\%$\pm$2.38\% & 97.90\%$\pm$0.25\% & 90.36\%$\pm$1.12\%   &  0.99$\pm$0.0021 \\
\bottomrule
\end{tabular}
    \caption{Stability of the empirical results considering both data split randomness and training randomness.}
    \label{tab:random}
\end{table*}

\section{PLM Backbone Comparison}
\label{app:plm}
In addition to Longformer, we also experiment with other PLM backbones such as BERT~\cite{bert}, RoBERTa~\cite{roberta}, and GPT2~\cite{gpt2}. The results of these experiments are shown in Table~\ref{tab:plm_comp}.
Firstly, the Longformer detector achieves the best performance in terms of both AvgRec and AUROC due to its ability to handle longer texts, while maintaining a small model size for efficient detection.
Secondly, increasing the model size improves detection performance for each backbone PLM.
Thirdly, masked language models (BERT, RoBERTa, and Longformer) outperform causal language models (GPT2).

\begin{table*}[t!]
\centering
\small
\begin{tabular}{cccccc}
    \toprule
     \textbf{PLM} & \textbf{\# Parameters}& \textbf{HumanRec}& \textbf{MachineRec} &\textbf{AvgRec}  & \textbf{AUROC}\\
     \midrule
     BERT-base & 110M & 67.11\%  & 98.34\%   &  82.72\% & 0.97 \\
     BERT-large & 336M  &  80.96\% & 93.27\%   &  87.12\%  & 0.96\\
     RoBERTa-base & 125M & 72.29\% & 95.28\%   &  83.78\% & 0.96\\
     RoBERTa-large & 355M  &  70.81\% & 98.38\%   &  84.59\% & 0.98 \\
     GPT2 & 117M & 57.42\% & 97.84\%   &  77.63\% & 0.96\\
     GPT2-medium & 345M & 69.94\% & 96.82\%   &  83.39\% & 0.96\\
     GPT2-large & 774M &  84.27\% & 96.67\%   &  90.47\% &  0.98\\
     Longformer & 149M & 82.80\% & 98.27\%   &  \textbf{90.53\%} & \textbf{0.99} \\  
\bottomrule
\end{tabular}
    \caption{Performance comparison of different PLM-based classifiers.}
    \label{tab:plm_comp}
\end{table*}

\section{Data Balance}
\label{app:balance}
Since the number of machine-generated texts is larger than that of human-written ones in the train set.
We investigate whether such an imbalance has an impact on the model performance.
Specifically, we randomly sample machine-generated texts to be the same quantity as human-written ones.
We experiment on the Longformer detector and present the results in Table~\ref{tab:data_balance}.
Despite the narrowed gap between HumanRec and MachineRec, we can observe that data balance has little influence on model performance in terms of AvgRec and AUROC.
In addition, the tendency of the Longformer detector to classify human-written texts as machine-generated ones still exists with a perfectly balanced training set.

\begin{table}[t!]
\centering
\small
\begin{tabular}{cccc}
    \toprule
     \textbf{HumanRec}& \textbf{MachineRec} &\textbf{AvgRec} & \textbf{AUROC} \\
     \midrule
     85.38\% & 92.95\% & 89.16\%   &  0.99 \\
\bottomrule
\end{tabular}
    \caption{Effects of data balance on detection performance (Longformer) under the \textit{Arbitrary-domains \& Arbitrary-models} setting.}
    \label{tab:data_balance}
\end{table}




\begin{table*}[t!]
\centering
\small
\begin{tabular}{l c c c c}
\toprule
\textbf{Methods} & \textbf{HumanRec} & \textbf{MachineRec} & \textbf{AvgRec} & \textbf{AUROC} \\
\midrule
\multicolumn{5}{c}{Unseen Domains \& Unseen Model}\\
\midrule
FastText & 71.78\% & 68.88\% & 70.33\% & 0.74 \\
GLTR & 16.79\% & 98.63\% & 57.71\% & 0.73 \\
Longformer & 52.50\% & 99.14\% & 75.82\% & 0.94 \\
Longformer\dag & 88.78\%\dag & 84.12\%\dag & 86.54\%\dag & 0.94 \\
\midrule
\multicolumn{5}{c}{Paraphrasing Attack}\\
\midrule
FastText & 71.78\% & 50.00\% & 60.89\% & 0.66 \\
GLTR & 16.79\% & 82.44\% & 49.61\% & 0.47 \\
Longformer & 52.16\% & 81.73\% & 66.94\% & 0.75 \\
Longformer\dag & 88.78\%\dag & 37.05\%\dag & 62.92\%\dag & 0.75 \\
\bottomrule
\end{tabular}
\caption{Detection performance on the two challenging test sets. `\dag' denotes the boundary is adjusted.}
\label{tab:ood_para_perform}
\end{table*}

\begin{table*}[t!]
\small
    \centering
    \begin{tabular}{cccc}
    \toprule
         \textbf{Data Source} & \textbf{Human-written} & \textbf{Machine-generated} & \textbf{All} \\
         \midrule
         Average Document Length & 232.02 & 279.99  & 263.87   \\
         Average Sentence Length &  18.90 & 18.80 &  18.83 \\
         Average $\#$ Sentences per Document & 13.48 & 15.33  & 14.71   \\
     \bottomrule
    \end{tabular}
    \caption{Length statistics for human-written and machine-generated samples.} 
    \label{tab:length}
\end{table*}

\begin{figure*}[t!]
\small
\centering
\includegraphics[width=0.7\linewidth]{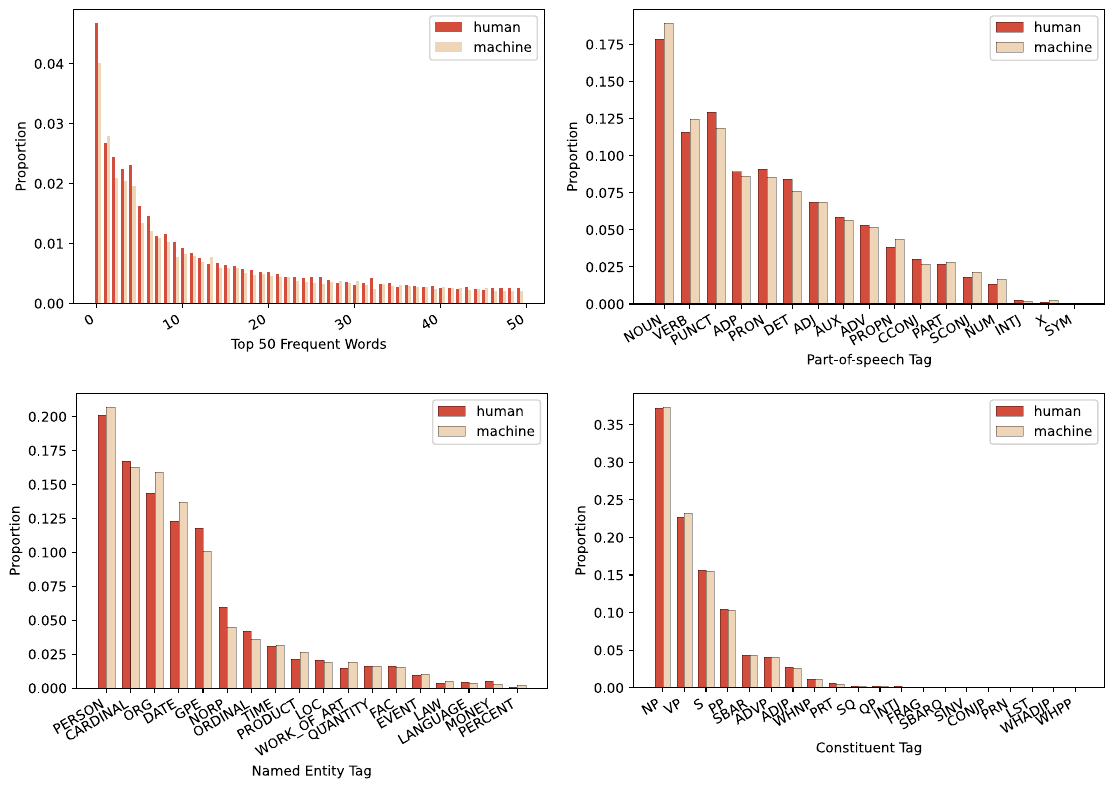}
\caption{
\label{fig:ling}
Linguistic statistics (word frequency distribution, part-of-speech distribution, named entity distribution and constituency distribution) for human-written and machine-generated samples.}
\end{figure*}

\begin{figure}[t!]
\centering
\includegraphics[width=0.8\linewidth]{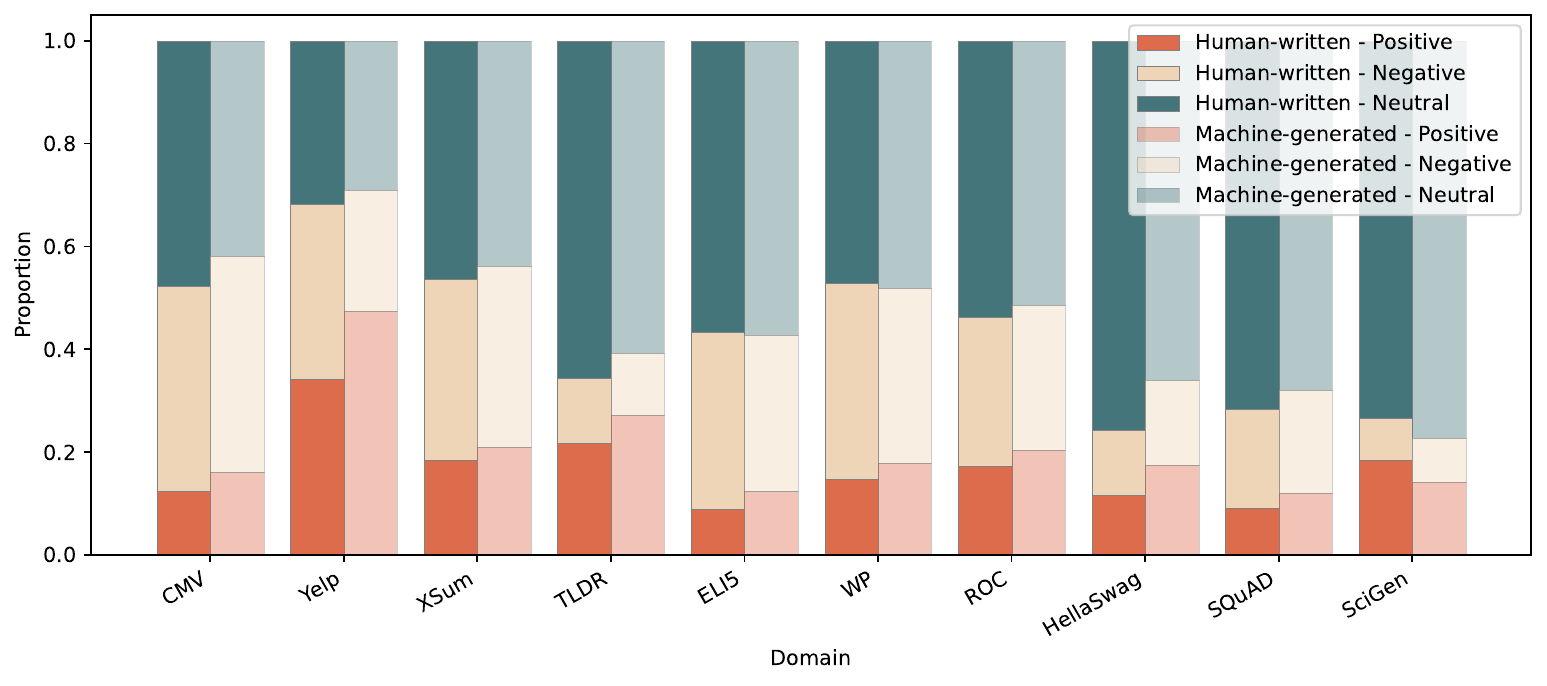}
    \caption{Sentiment polarity.}
    \label{fig:sentiment}
\end{figure}

\begin{figure}[t!]
\centering
\includegraphics[width=0.8\linewidth]{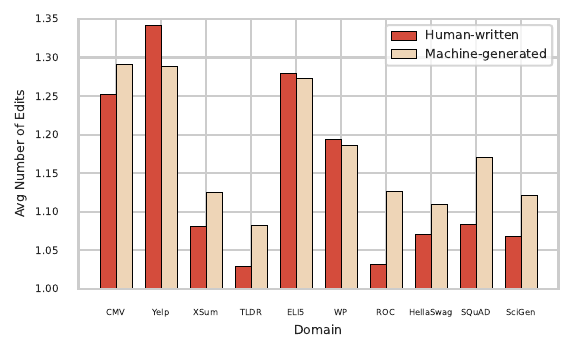}
    \caption{Grammar formality. A lower number of edits indicates better grammar formality.}
    \label{fig:grammar}
\end{figure}

\section{Detection Performance on the Two Challenging Test Sets}
\label{app:ood_para_perform}
The detection performance of all methods on the two challenging test sets, i.e., Unseen Domains \& Unseen Model and Paraphrase Attack, is shown in Table~\ref{tab:ood_para_perform}.
Detect-GPT is not included due to its reliance on the white-box detection setting.
We can observe that all methods suffer severe performance degradation in terms of AUROC, indicating weakness in detecting machine-paraphrased texts.

\section{Text Characteristics}
\label{app:stat}
We first explore to find potential surface patterns that can help discriminate between human-written texts and machine-generated ones.
The length statistics are shown in Table \ref{tab:length}.
As can be seen from the table, although we do not exert explicit length control over the model generation, the average length of machine-generated texts is marginally longer than that of human-written.

\paragraph{Linguistic Pattern.} We further use Stanza, a linguistics analysis tool \citep{stanza}, to gain a more systematic understanding of the linguistic components in both sources, with results shown in Figure \ref{fig:ling}.
We can observe that texts from both sources share similar distributions under various linguistic scales, such as word frequency, part-of-speech frequency, named-entity frequency, and constituent frequency.
In other words, there is no significant linguistic difference between the text sources (human-written versus machine-generated) that can assist the classifier in differentiating them in a wild setting.

In addition, we explore whether there are differences between human-written and machine-generated texts in other characteristics (such as sentiment polarity and grammar formality) when considering diverse writing tasks and various text-generating LLMs.

\paragraph{Sentiment Polarity.}
We use an off-the-shelf sentiment classifier \cite{sentp} trained on 198M tweets for sentiment analysis to analyze the sentiment polarity of both texts, with results shown in Figure \ref{fig:sentiment}.
As suggested by \citet{chatgpt_detect}, ChatGPT expresses more neutral sentiments than humans.
In a large-scale setting that considers various domains and LLMs, however, there is no clear distinction between human-written and machine-generated texts in terms of sentiment polarity. Notably, LLMs generally generate more positive texts, especially when creating reviews or comments (Yelp).

\paragraph{Grammatical Formality.}
We use an off-the-shelf grammar error correction model \cite{yue_gec} to evaluate the grammar formality of human-written and machine-generated texts.
We adopt the average number of edits to quantify grammar formality.
As shown in Figure \ref{fig:grammar}, machine-generated texts are equally or even more grammatical in domains (CMV, Yelp, ELI5, and WP) where texts are less formal (reviews or posts on forums).
In formal domains such as XSum (news articles), SQuAD (Wikipedia documents), and SciXGen (scientific writings), human-written texts exhibit better grammatical formality.




\end{document}